\documentclass[%
 preprint,
superscriptaddress,
 amsmath,amssymb,
 aps,
 pre,
table, 
xcdraw
]{revtex4-1}

\usepackage{graphicx}% Include figure files
\usepackage{dcolumn}% Align table columns on decimal point
\usepackage{bm}% bold math
\usepackage{xcolor}
\usepackage[bf]{subfigure}
\usepackage{tcolorbox}
\usepackage{multirow}
\usepackage{array}

\usepackage{color}
\definecolor{redcolor}{rgb}{0,0.,0.}
\def\Red#1{{\color{redcolor} #1}}
\definecolor{bluecolor}{rgb}{0,0.,1}

\begin{document}

\preprint{APS/123-QED}

\title{Text Characterization Based on Recurrence Networks}
\author{Bárbara C. e Souza}
\affiliation{Institute of Mathematics and Computer Sciences,
University of S\~ao Paulo, PO Box 369,
13560-970, S\~ao Carlos, SP, Brazil 
}

\author{Filipi N. Silva}
\affiliation{Indiana University Network Science Institute, Bloomington, IN, USA
}

\author{Henrique F. de Arruda}
\affiliation{
 S\~ao Carlos Institute of Physics,
University of S\~ao Paulo, PO Box 369,
13560-970, S\~ao Carlos, SP, Brazil
}
\affiliation{ISI Foundation, Via Chisola 5, 10126 Turin, Italy}

\author{Giovana Daniele da Silva}
\affiliation{Institute of Mathematics and Computer Sciences,
University of S\~ao Paulo, PO Box 369,
13560-970, S\~ao Carlos, SP, Brazil 
}

\author{Luciano da F. Costa}
\affiliation{
 S\~ao Carlos Institute of Physics,
University of S\~ao Paulo, PO Box 369,
13560-970, S\~ao Carlos, SP, Brazil
}

\author{Diego R. Amancio}
\affiliation{Institute of Mathematics and Computer Sciences,
University of S\~ao Paulo, PO Box 369,
13560-970, S\~ao Carlos, SP, Brazil 
}

\date{\today}% It is always \today, today,
             %  but any date may be explicitly specified

\begin{abstract}
Several complex systems are characterized by presenting intricate characteristics taking place at several scales of time and space. These multiscale characterizations are used in various applications, including better understanding diseases, characterizing transportation systems, and comparison between cities, among others. In particular, texts are also characterized by a hierarchical structure that can be approached by using multi-scale concepts and methods.  The multiscale properties of texts constitute a subject worth further investigation.  In addition, more effective approaches to text characterization and analysis can be obtained by emphasizing words with potentially more informational content. The present work aims at developing these possibilities while focusing on mesoscopic representations of networks. More specifically, we adopt an extension to the mesoscopic approach to represent text narratives, in which only the recurrent relationships among tagged parts of speech (subject, verb and direct object) are considered to establish connections among sequential pieces of text (e.g., paragraphs). The characterization of the texts was then achieved by considering scale-dependent complementary methods: accessibility, symmetry and recurrence signatures.  In order to evaluate the potential of these concepts and methods, we approached the problem of distinguishing between literary genres (fiction and non-fiction).  A set of 300 books organized into the two genres was considered and were compared by using the aforementioned approaches. All the methods were capable of differentiating to some extent between the two genres.  The accessibility and symmetry reflected the narrative asymmetries, while the recurrence signature provided a more direct indication about the non-sequential semantic connections taking place along the narrative.
\end{abstract}

\maketitle

\section{\label{sec:introduction}Introduction}

Several structures and dynamics in the natural, as well as in artificial, worlds involve several \emph{scales} regarding space, time, etc.  For instance, when observed from further away (large spatial scale), a forest will appear mostly homogeneous, with some possible  varying patches related to different plant species or terrain and weather-related effects. However, when observed at a closer range (mesoscopic scale), the same forest will appear substantially distinct as we observe branches and leaves.  At an even smaller spatial scale (microscopic), we will start perceiving the leaves venation, details on the surface of the stems, etc.  

Though, in principle, there is no limit to the range of possible scales in which the same structure or phenomenon can be observed and analyzed, it is reasonable to focus interest on the scales more closely related to the objectives of a given research.  In case one is interested, let's say, in studying the sunlight incidence and shades on the leaves of a forest, the mesoscopic and macroscopic scales would probably be more important.  However, as soon as interest is shifted to the effect of light absorption on the growth of the trees, more microscopic scales will need to be taken into account.

The interesting association of structural and dynamical properties with several types of scales can be observed almost without exception in the natural world as well as in other realms, including human language.  As with other living beings, along their long history, human beings need to develop effective abstractions to represent the most relevant aspects of their environment, not only for self-referencing, but mostly for communicating between individuals (e.g.~\cite{CostaModeling}).

The above mentioned representation of the real world into symbols that could be orally communicated, and later consolidated as written text, involved an interesting trade-off between specificity and generality, as well as between abstraction and detailed description.  While a more generic and abstract representation of some concepts may be enough in some cases, as soon as some specific aspects assume increased importance, new concepts need to be defined and incorporated. For instance, we can do with oranges to represent all possible of these fruits, but as soon as we are interested in their production in a specific terrain and climate, the overall category of oranges will need to be sub-partitioned into smaller subsets up to the level of species and subspecies.  

Interestingly, the successive partitioning of more generic, abstract concepts into new concepts and  subgroups establishes a \emph{hierarchy} of representations.  As studied systematically in areas as pattern recognition and artificial intelligence, hierarchies are inherently associated with specific scales.   Hence, we have that human language is an intrinsically multi-scale system, in which the levels of generality, detail, abstraction and specificity vary according to specific situations and demands while consolidating knowledge into written text, or while orally communicating between individuals.

While texts have been frequently studied from the perspective of word adjacency or proximity (e.g.~\cite{marinho2016authorship,santos-etal-2017-enriching}), the above discussed aspects of human language motivated more systematic approaches capable of taking into account not only smaller scales (e.g.~related to the composition of words), but also mesoscopic and macroscopic scales~\cite{ferraz2018representation}. Thus, in addition to considering the more local interrelationship between words (e.g.~by adjacency or proximity), it becomes important to systematically  approach texts in terms of sentences, paragraphs, sections, chapters, and even book collections and whole libraries related to specific themes or epochs.

In~\cite{amancio2015complex}, the authors used a networked representation formed by the co-occurrence of words to address the problem of identifying authors style. Differently from the traditional approach based on the frequency of words, the authors introduced a hybrid approach taking into account two main factors: (i) the frequency of words; and (ii) the topological measurements of complex networks. Interestingly, the authors found that frequency- and topological-based approaches complement each other, since both strategies yield provided useful, complementary information to identify authors' styles. Similar word co-occurrence networks have also been used in additional contexts~\cite{garg2018identifying,fatima2021dasentimental,stella2021mapping}. While this approach capture some of the temporal/spatial narrative, the word level approach can only extract syntactical and stylistic features of texts~\cite{de2019paragraph}. In addition, the traditional co-occurrence approach does not link similar words. %Some approaches to address this particular issue has been recently reported in~\cite{}. 

In~\cite{correa2020semantic}, the semantic flow of texts was studied. In the proposed approach, nodes are sentences and edges are established by taking into account the semantic similarity of the respective nodes. The authors found that the transition between semantical clusters can be used to discriminate between distinct styles. In particular, the semantical flow allowed discriminating philosophy from investigative books with an accuracy larger than 92\%. In~\cite{de2019paragraph} the authors analyzed networks formed via paragraph semantic similarity. The authors found that this type of representation complements traditional word co-occurrence networks because paragraph networks can grasp semantic features of texts. 
While the proposed paragraph network is able to go beyond syntax/style, the narrative temporal aspect is not taken into account since paragraph order is not taken into account while creating the network. 

In the present work, we aim at studying texts, more specifically books from the Gutenberg project~\cite{gerlach2020standardized}, from the mesoscopic perspective of linear sequences along the text, as well as the accessibility and symmetry~\cite{SILVA201661,TRAVENCOLO200889} of texts when represented as mesoscopic networks.  

The former approach involves treating the text as a linear sequence of paragraphs while identifying tf-idf~\cite{manning1999foundations} cosine similarity between the obtained paragraphs along the whole sequence. A recurrence network~\cite{donner2010recurrence} is respectively obtained, in which the interconnections indicate short to long range relationships along the text narrative.  These relationships, as gauged by the co-occurrence of words, are likely to indicate recurrent situations in space, time or subject along the narrative.  For instance, a location may recur along the text, giving rise to several respective longer range links.  This same effect can occur with characters. 

\Red{Different from other approaches that only captures local (or co-occurrence) similarity~\cite{waumans2015topology,amancio2016network}, our approach is able to capture long-range references in the text narrative. The limitation with approaches based on word adjacency concerns the fact that single words are typically not enough for characterizing a well-defined context.  By employing paragraphs (or even larger portions of the text), it becomes possible to establish correspondences between paragraphs referring to the same context, such as a situation, place, character, etc.  Given that these paragraphs can be far away one another (e.g.~in different chapters), \emph{long range} connections can be obtained.}

The other approach integrated in the current analysis concerns the estimation of the accessibility and symmetry of each of the nodes in paragraph-based networks.  The
accessibility was proposed as a means to quantify how effectively, according to a specific dynamics taking place on a network, other nodes can be accessed by a given network node~\cite{travenccolo2009border}. Interestingly, this measurement intrinsically incorporates means for investigating multi-scale relationships in the analysed data, which is achieved by varying the order of the considered neighborhood~\cite{TRAVENCOLO200889}.

Several interesting results are reported and discussed in the present approach, including the identification of the potential of the accessibility, symmetry, and recurrence signatures for distinguishing between distinct literary genres, suggesting that these scale-dependent measurements are capable of quantifying the heterogeneity of the narrative.  Presenting noticeable correlation with the accessibility and symmetry, the recurrence signatures can be more directly related to the properties of the narratives.  

The current work starts by presenting the basic concepts and methods adopted, and proceeds by presenting the application of the described measurements respectively to literary genre discrimination.

\section{\label{sec:model} Materials and methods}

\begin{figure*}[htbp]
    \centering
    \includegraphics[scale=0.5]{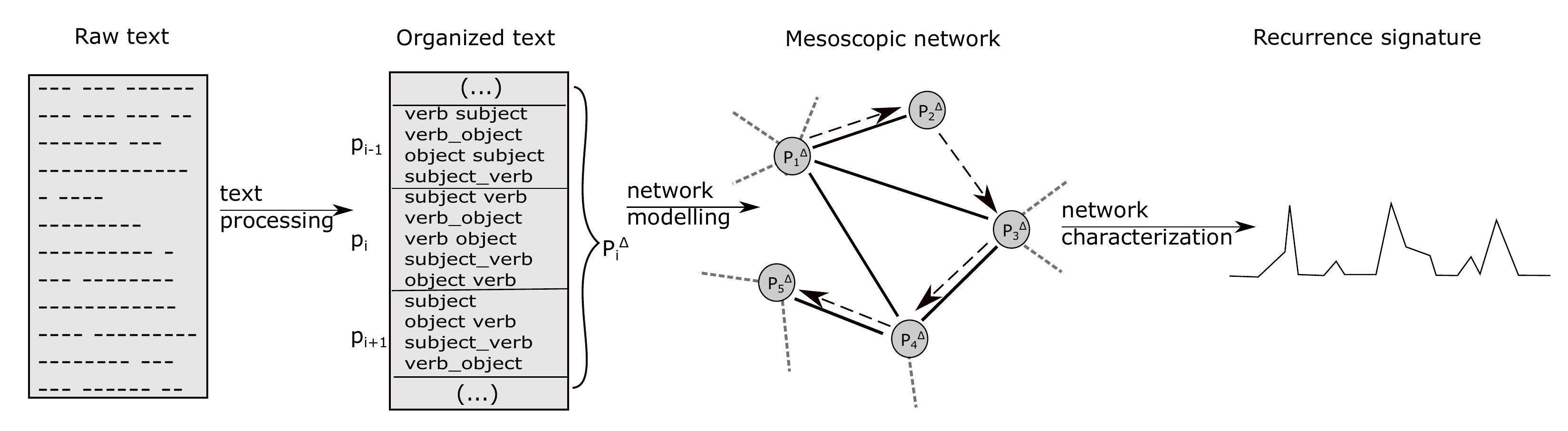}
    \caption{Diagram of the execution pipeline of the methodology proposed. The pre-processing of the raw text yields the organized text (O), which is then used to generate the mesoscopic network. Finally, measurements such as the Recurrence Signatures proposed can be extracted from the network to characterize it.}
    \label{fig:pipeline}
\end{figure*}

This section describes the procedure employed to obtain (and characterize) networks from any text, which includes books and \Red{any other documents structured with paragraphs.} The adopted pipeline is illustrated in Figure \ref{fig:pipeline}. The procedure is described as follows:

\begin{enumerate}

    \item \emph{Text processing}: in this step we are interested in analyzing the relationship between specific words. For example, we aim at identifying recurrent behavior by the same subject. For this reason, a syntactic parsing is applied in order to identify such relevant words. In this step, words conveying low semantic meaning \Red{(such as prepositions and articles, usually known as stopwords) are also disregarded, since they are not relevant to characterize the narrative semantic flow, or, in other words, the story constructed within the text.}
    
    \item \emph{Network modelling}: here, a semantic representation of the narrative flow is created, where nodes represent a sequence of paragraphs and edges are established according to the semantic similarity between the nodes.  
    
    \item \emph{Network characterization}: the characterization of the generated networks is performed via network measurements. We also proposed a signature characterization based on the recurrence of semantic context along the narrative.

\end{enumerate}

The computational expenses associated to the proposed methodology is detailed in Appendix A. 
Additional aspects about the adopted procedure are described
as follows.

\subsection{Dataset and genre classification} \label{sec:dataset}

The dataset used in this project consisted of 300 different books. They were all retrieved from a random selection of the Project Gutenberg \cite{gutenberg}. The selection considered only books written in the English language written between 1000 and 2000 paragraphs. This selection was not specific to any other aspect of the books, i.e. author, publication date, literary genre, etc.

The Gutenberg Project provides open access to books written in several different languages. For each book, in addition to the full content, additional metadata are provided. This includes author, illustrator, title, language and associated subjects \Red{(which will be referred to as literary genres in this document)}. In this study, we retrieved, for each book, its content in raw text format together with its set of genres.

An important aspect of this dataset is the lack of a well defined classification for a book's genre. Since Gutenberg Project provides a set of genres without any distinction of importance, there is no straightforward way of assigning one specific genre to each book. This issue is also complicated by the fact that there is not a single granularity for the book's genres other than a few general ones such as \emph{PR (Language and Literature: English literature)} or \emph{PZ (Language and Literature: Juvenile belles lettres)}, as well as particularly specific ones, such as \emph{Scarecrow Fictitious character from Baum} and \textit{National Research and Education Network Computer network}. Another problem is that a single book can be labeled with several of the non-informative labels.

\begin{figure}[htb]
    \centering
    \includegraphics[scale=1.5]{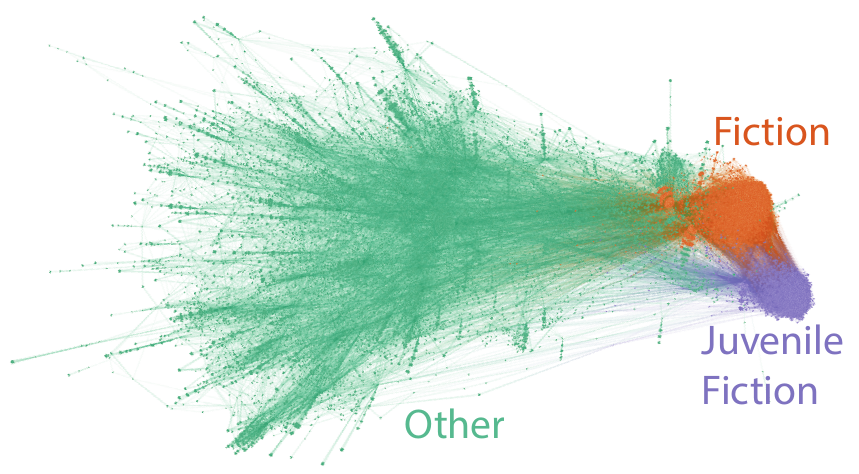}
    \caption{Visualization of the projection network onto genres for the whole Gutenberg dataset, colored by the community detected. Generated using the \emph{Networks3d} software~\cite{silva2016using}.}
    \label{fig:communities}
\end{figure}

In order to better understand the relationship between the labels provided by Gutenberg Project, we constructed a bipartite network between books and genres, linking a book with every one of its literary genres. Next, a projection onto the genres was created from the bipartite graph. Therefore, \Red{each node represents} an existing genre, and a connection is implemented between two genres whenever there is at least one book containing both. Finally, a network with approximately $40,000$ nodes was obtained. A respective visualization is shown in Figure \ref{fig:communities}. 

\Red{Since it is impracticable to deal with a set of that many labels, we needed to increase the granularity level of the genre set. In order to cluster the set of genres into these broader labels, the Louvain community detection algorithm \cite{louvain} was applied.} Three main communities were found. After an empirical analysis, we found that the first one, in orange, comprises mostly genres related to adult fiction, while the second one, in purple, consisted mostly of juvenile fiction. Lastly, the third and most sparse one, in green, contains all the other possible genres in the Gutenberg Project, including history, art, bibliographies, etc.

Finally, we can refer to these communities in order to define a label for a given book. Since every literary genre can now be associated with a single community in the network, it can also translate to a single label. Therefore, one can assign a label to any book by choosing the community that comprises the majority of that book's genres.

\subsection{From texts to networks} 

When tackling any sort of problem from a real life system, it is essential to consider an appropriate scale to develop a solution that fits the phenomena being studied. Consider, for illustration purposes, any natural environment: there are numerous problems that demand an overall view of the macro aspects of that system. Take weather forecast as an example: to analyze and predict the weather of that specific system, one must consider features such as location, vegetation, humidity, and others, while the micro aspects such as each individual animal in the fauna or plant in the flora are not as important for the problem in question. It is very common to observe problems that focus either on the macro or the micro scale of the system, however, there is still a wide scope of intermediate scales in between these two extremes that are yet to be explored.

In the context of text analysis, this issue is also present: while there are many different works that focus on the microscopic scale of the texts (e.g. word adjacency networks~\cite{stella2020multiplex,stella2019forma,mehri2012complex}) or on the macroscopic scale (e.g. citation networks~\cite{tang2021trends}), there are still substantially fewer works that are placed anywhere between the two extremes. 

Only more recently, techniques have been more systematically considered in order to create networks from documents, while taking into account their mesoscopic structure~\cite{de2019paragraph,correa2020semantic}. In \cite{DEARRUDA2018110}, for example, the authors apply image analysis techniques to network visualizations in order to extract topological features of the mesoscopic networks to tackle the authorship recognition problem. Still in the context of this problem, in \cite{DBLP:journals/corr/MarinhoALCA17}, the authors make use of the ability of the mesoscopic network to capture a narrative's story flow and, hence, the author's ``calligraphy". Finally, in \cite{ferraz2018representation}, the mesoscopic networks are used to study and analyze texts while considering the overall narrative and its unfolding along time.

As an extension of the concepts presented in the previously mentioned works, here we focus on these mesoscopic aspects of the texts, often overlooked by more traditional approaches, to propose a new technique to model networks from texts. By looking at texts from a mesoscopic perspective, it is possible to grasp the semantic context of a narrative, which can potentially be used to tackle a variety of different problems, such as genre classification and authorship recognition. This methodology, summarized in the pipeline shown in Fig. \ref{fig:pipeline}, will be explained in detail hereon. 

Differently from other typical approaches, here we do not start with punctuation or stopwords (words conveying little contextual meaning, e.g. prepositions and articles) removal, since those are important for the employment and performance of the syntactical dependency analysis method. Hence, the only clean up performed at this point is the removal of underscores ("\_") and chapter markers, since they can introduce noise to the syntactic analysis that will be performed next. With the same goal, a co-reference resolution technique is applied~\cite{manning2014stanford}. Given that co-references are expressions that refer to a previous mentioned entity in the text, (e.g. pronouns), this technique ensures that there is a minimum occurrence of multiple terms referring to the same entity. 

After this pre-processing step, syntactic analysis is employed, where each paragraph is reduced to a set of tokens with specific syntactic roles: either \textit{subject}, \textit{verb} or \textit{direct object}. This role selection was chosen with the intention of getting only the tokens that provide the most contextual meaning to the sentences where they are found, while excluding any possible noise. This strategy was based on the fact that, in order to understand the semantics of something, it is usually necessary to answer the questions \textit{what is happening} and \textit{what or who is doing it}. To obtain an answer to the latter, we retrieve the subject in the sentence, which is the entity responsible for the activity happening. Secondly, when recovering the verb and its direct object, we are referring to the action occurring and, therefore, answering the first inquiry. This approach is also supported by the fact that the incorporation of linguistic knowledge can contribute to text summarization \cite{mutlu2020candidate}, and, therefore, to capture the contextual meaning of texts by enhancing informativeness. In this work we used the CoreNLP parser~\cite{manning2014stanford}. 

At last, each of these tokens is normalized to its canonical form by using a lemmatization technique \cite{10.5555/311445}, resulting in the disregard of inflections in verbal tense, number, case or gender. Therefore, the sentence "thought Alice to herself", for example, will first be transformed to "thought Alice to Alice", then to "thought Alice" and, finally, to "think Alice". By that, the semantics of the sentence is actually summarized in the final set of tokens, where it is possible to answer \textit{what} is happening (by the verb \textit{think}) and \textit{who} is taking that action (by the subject \textit{Alice}). 

\begin{figure*}[htb]
    \centering
    \includegraphics[scale=0.6]{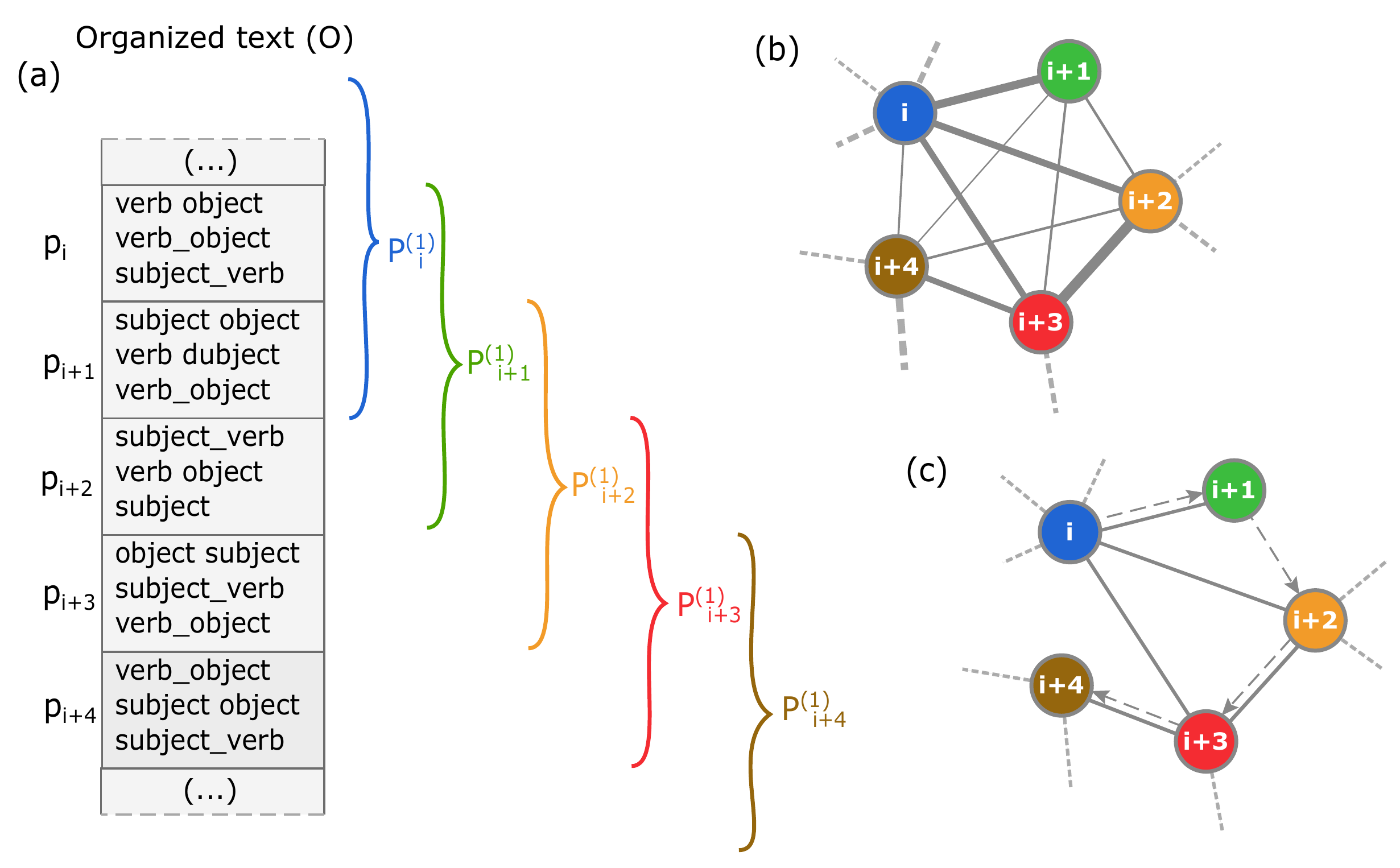}
    \caption{Illustration of the presented methodology on how to construct the mesoscopic network based on the organized text $O$ (a). Initially, there is a fully connected weighted network, where the edge weights are the cosine similarity calculated between both nodes and are illustrated in (b) by the width of the lines. Next, only the $|V| \times T$ strongest connections are maintained, to ensure that the average degree of the graph is now equal to $T$. Finally, the sequence edges are added between consecutive vertices, as illustrated by the dashed arrows in (c).}
    \label{fig:methodology}
\end{figure*}

After all this processing, one gets the final organized text \(O\), as it will be called from here on. $O$ is a sequence of paragraphs \(O = (p_0, p_1, p_2...)\), each comprising a sequence of words \(p_i = (w_{i0}, w_{i1}, w_{i2}...)\). Next, to build the mesoscopic network, in contrast to the word adjacency model, a sequence of paragraphs \(p_{i-\Delta}, p_{i-\Delta+1}, ...p_i, p_{i+1}, ...p_{i+\Delta}\) is mapped into a node $i$, obtaining $P_i^{(\Delta)}$. In Figure \ref{fig:methodology} we illustrate this process of getting the paragraph windows $P_{i}^{(\Delta)}$, for their respective nodes in the mesoscopic network, considering $\Delta = 1$.

\Red{Next, to create the edges, we employ the techniques \textit{bag-of-words} followed by the \(\operatorname{tf-idf}\). At first, we use the  bag-of-words to represent the text from now on, since only the tokens are important now and not the actual text structure. Then, we apply the TF-IDF technique considering a text as a collection of documents (in this case, paragraphs) and calculate the score between each of them. Hence, the set of documents $D$ is $O$, where each document $d$ is a paragraph window $P_i^{(\Delta)}$ (the text comprised by one node in the network); and $w$ is each one of the words of a paragraph. Finally, cosine similarity is used to calculate $sim(P_A, P_B)$ from the TF-IDF scores, between nodes of indexes $A$ and $B$, where $|A-B| > \Delta$.} This second restriction is due to the fact that, otherwise, the two paragraph windows share at least one paragraph and, therefore, the similarity computed between them would be biased.   

As a result, a fully connected network is created (see Figure \ref{fig:methodology}(b)), in which the edge weights correspond to the similarity $sim(P_A, P_B)$ normalized between 0 and 1 among each pair of nodes. The final mesoscopic network is obtained by pruning the weakest connections until the average degree of the network reaches a specified threshold \(T\). After this procedure, edge weights are ignored, resulting in an unweighted network (see Figure \ref{fig:methodology}(c)) with a fixed average degree greater than  0. All of these edges will be referred to as \textit{similarity edges} hereafter. In addition, \(|O| - 1\) edges are inserted linking nodes that represent adjacent paragraphs, that is, \(P^\Delta_1\) will be linked to \(P^\Delta_2\), \(P^\Delta_2\) to \(P^\Delta_3\) and so on, as illustrated in Figure \ref{fig:methodology}(c) by the dashed edges. These edges are marked as \textit{sequence edges}. 
Such edges will guarantee that the network is a connected component. In addition, sequence edges provide a temporal narrative perspective, as it happens in traditional word adjacency networks~\cite{amancio2015concentric,kulig2015modeling,segarra2015authorship}. 

\subsection{Network topology measurements}

To understand and discriminate a network's topology, it is essential to be able to extract measurements from the graph that reflect the desired properties. The accessibility measure \cite{TRAVENCOLO200889, Viana}, for example, was proposed to quantify the number of effectively accessible nodes given an established distance while respecting specific dynamics. In that manner, the accessibility can measure how peripheral or central a node actually is, which can be useful to reflect the network topology as a whole when considering all of its nodes individually. In addition, accessibility-based measurements can also identify relevant words in texts, being thus useful to detect keywords and discriminate authors~\cite{amancio2012extractive,amancio2015concentric}.

The definition of this measurement is based upon the concept of random walks (or any other dynamics) and concentric levels. In a random walk, there is an agent that moves between the nodes of a network through its edges. One variation of this definition is the self-avoiding random walk, in which the agent cannot go through the same node more than once. This kind of walk is the one used to define the accessibility measure. Besides, the concentric level $h$ of a node $i$ is defined as the set of nodes that are at a distance $h$ when departing from $i$. Moreover, it is possible to define the probability vector $p_i(h) = \{p_1^{(h)}, p_1^{(h)}, ... , p_{N_i(h)}^{(h)}\}$ of reaching each one of the $N_i(h)$ neighbors of $i$ in its concentric level $h$, when considering a self avoiding random walk. Hence, the accessibility value $k$ for a node $i$ and a concentric level $h$ can be calculated as:
\begin{equation}
    k_i(h) = \exp \Big{(} -\sum_j {p_j^{(h)} \log p_j^{(h)}} \Big{)}.
\end{equation}

\begin{figure*}[htb]
    \centering
    \includegraphics[scale=0.7]{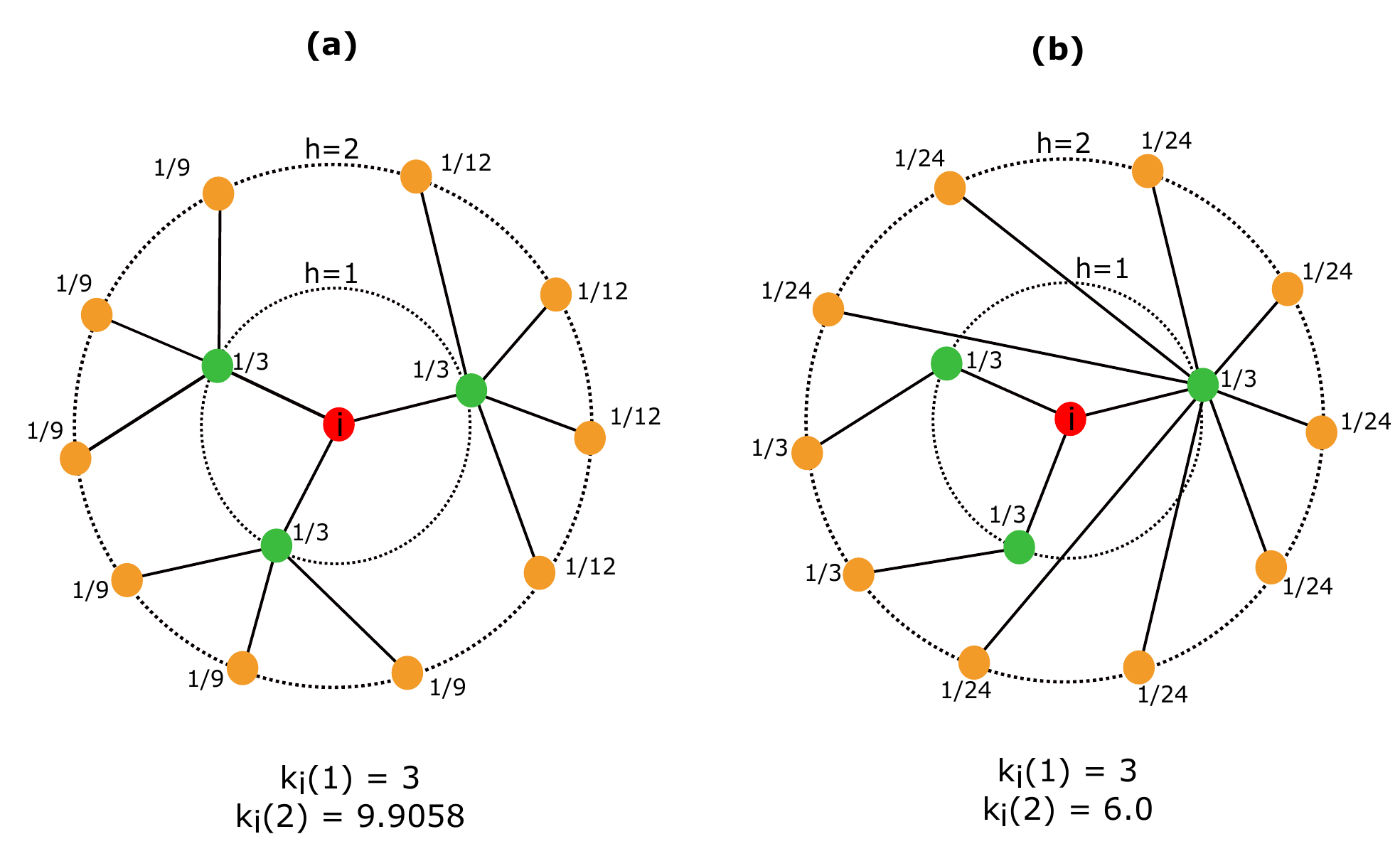}
    \caption{Examples of the accessibility calculation for node i (in red). By definition, for h=1, the accessibility value is the actual concentric degree and is trivially obtained for both networks. In (a), given the network topology, the nodes in the second concentric level have nearly the same probability of being reached, hence the high value obtained for the accessibility measure (considering that the maximum would be 10). In (b), since the discrepancy of the probabilities is greater, the value obtained for accessibility is smaller.}
    \label{fig:accessibility}
\end{figure*}

Differently from the node degree, the accessibility measurement considers how many nodes can be effectively accessed, given the  probability vector $p_i(h)$. In Figure \ref{fig:accessibility}, we show two examples of the accessibility calculation for the node $i$, in red. Considering the first concentric level $h=1$, in green, the calculation is trivial, since, by definition, it is actually the concentric degree. However, for the second level $h=2$, it is already possible to note a difference between the two values. In Figure \ref{fig:accessibility}(a), the probability vector of reaching each neighbor of $i$ is more uniform, which reflects in the high value of accessibility obtained, approximately 9.9. The magnitude of this value can be confirmed by considering the theoretical maximum by definition is 10, the number of nodes in that concentric level. In contrast, for Figure \ref{fig:accessibility}(b), one can see that the accessibility value for $h=2$ is considerably smaller. That can also be explained by the probability vector of reaching the neighbors of $i$, which shows a greater discrepancy between each of its values.

\begin{figure*}[htb]
    \centering
    \includegraphics[scale=0.7]{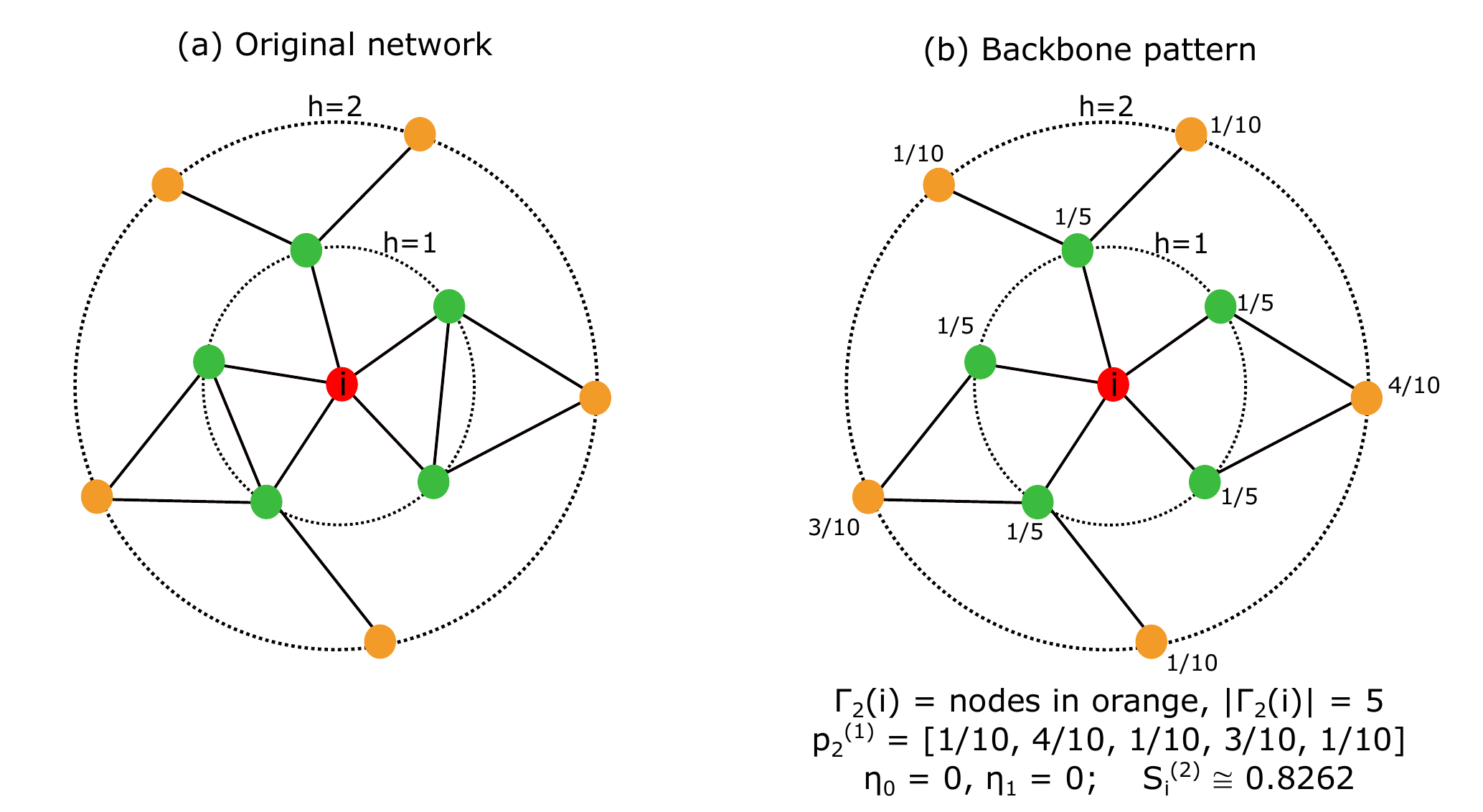}
    \caption{Example of the symmetry calculation for node i (in red) and level $h=2$. In (a), the original network is shown. In (b), one can see the backbone pattern for the original network, the one that will be used to calculate the symmetry value. We also have the parameter values used for the calculation and the final symmetry value.}
    \label{fig:symmetry}
\end{figure*}

From the concept of accessibility, concentric levels and probability vectors when considering random walks, the authors in \cite{SILVA201661} also derived the definition of concentric symmetry, which will also be useful for this work. The backbone pattern of the concentric symmetry, which is the one considered hereafter, is created by removing the edges between nodes in the same concentric level, as shown in Figure \ref{fig:symmetry}. In their work, the authors also follow up defining the merged pattern, that consists of merging together the nodes that were connected by these removed edges. However, the usage of the merged pattern for our classifications yielded no relevant improvement when compared to the backbone pattern, and, therefore, we will only consider the latter for this work, for simplicity sake.

Finally, the symmetry value can be calculated using the equation: 
\begin{equation}
    S_i^{(h)} = \frac{\exp{(-\sum_{j \in \Gamma_h(i)}(p_{ij}^{(h)} \times \log p_{ij}^{(h)}))}}{|\Gamma_h(i)| + \sum_{r=0}^{h-1}\eta_r},
\end{equation}
where \(\eta_r\) is the number of nodes that are not connected to the next concentric level (\(h+1\)), \(\Gamma_h(i)\) is the set of neighbors of \(i\) that also belong to level \(h\) and \(p_{i}^{(h)}\) is the probability vector of node $i$ considering the concentric level $h$. The specifications of these parameters can also be found in Figure \ref{fig:symmetry}, alongside with the symmetry value calculated for the network in question.

Because of how they work and what specific network properties they can capture, the accessibility and symmetry measures can be specially interesting for this work. These measurements are able to capture the heterogeneity of the network in question, by considering the existence of \textit{jumps} from one region of the network to another. Moreover, they are considered multi-scale, that is, it is possible to adapt the path length considered to reach further or closer regions of the network, which is also important for the text characterization problem in question.

\subsection{Network recurrence signature}
Here, we propose a measurement that is intended to capture the overall structure of the network, assuming its visualization using the force-directed nodes placement technique described in \cite{silva2016using}. With this approach, each network forms a continuous line, representing the story told along the book. This line can be bent to approximate two different points of the book, indicating that the content of those parts are similar to each other. Therefore, the goal of this new proposed measurement is to capture the overall network structure constructed from these bents, when they happen throughout the story line, how long they are and how often they happen.

\begin{figure}[htbp]
    \centering
    \includegraphics[scale=0.45]{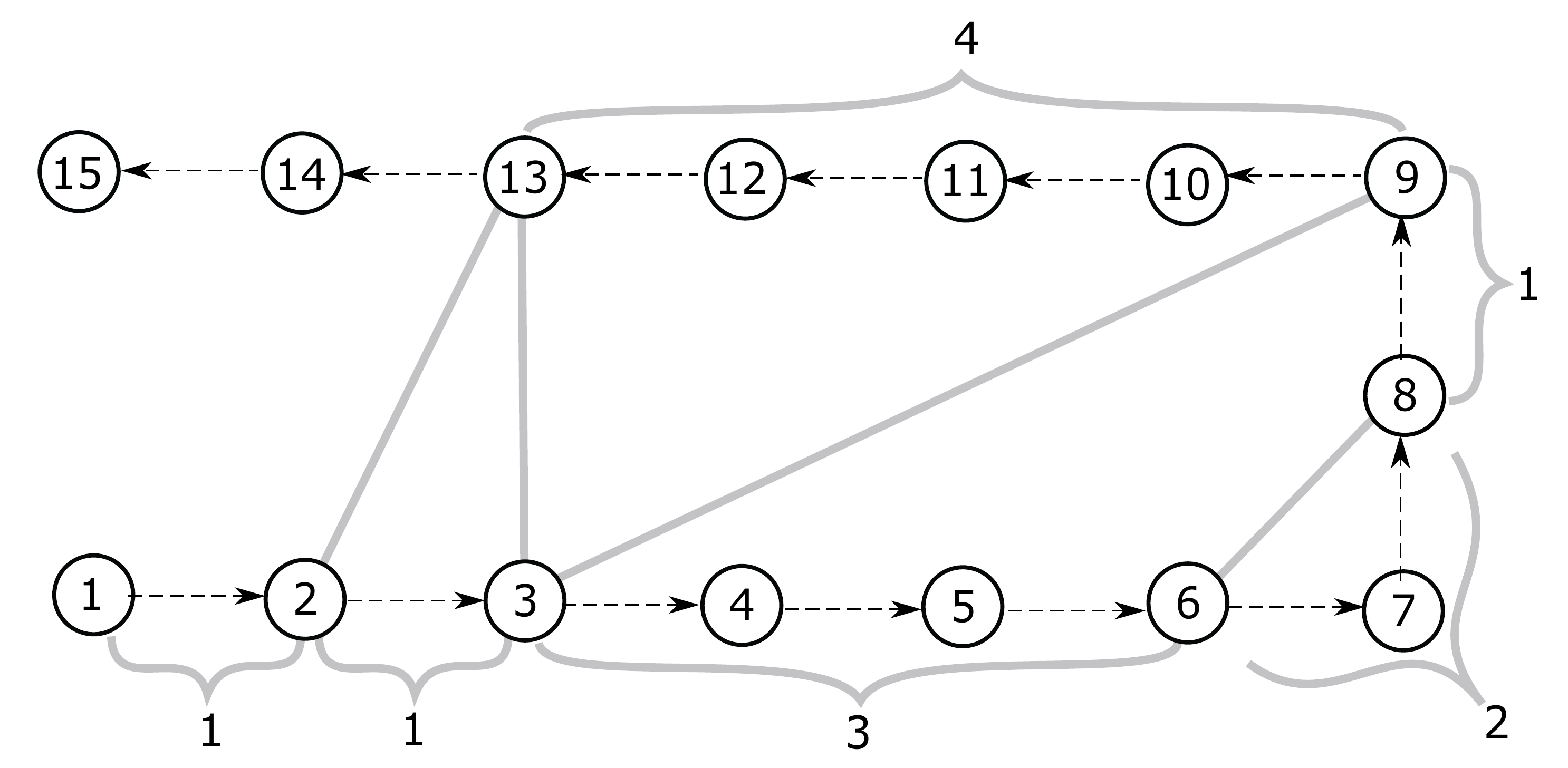}
    \caption{Visual representation of the proposed recurrence signature \(RS\). Here, initially the counter is set to 0, and the analysis starts from vertex 1. Since it does not have any similarity edges, nothing is done. After jumping to node 2, we increment the counter by 1 and evaluate it. Since there is a similarity edge there, the value of counter is appended to the array (which is now [1]) and the counter is set back to 0. After getting to vertex 3, the counter is incremented again and, since there is another similarity edge there, another 1 is added to the array (getting [1, 1]), the counter is set back to 0 and the iteration continues to the next vertex. The algorithm stops when reaching node 15, the end of the graph, where the final signature \(RS = [1, 1, 3, 2, 1, 4]\) is retrieved. 
    }
    \label{fig:series1}
\end{figure}

With that in mind, a series that can be extracted from a network is proposed, which will be referred to, hereafter, as Recurrence Signature (RS). This measure, shown in Fig. \ref{fig:series1}, captures when any bent happens in the network, that is, when there is a similarity edge connected to the node in question. It is important to notice that this cross reference can refer to the past or to the future of the story, and it can have any length greater than zero, which is completely disregarded, since the only important aspect is the mere existence of a cross reference. The signature is constructed by following the algorithm below: 
\begin{enumerate}
    \item Initialize a counter with 0;
    \item Initialize an empty array, which represents the network signature;
    \item For each of the network vertices, in order, check if there's a similarity edge in that vertex:
    \begin{itemize}
        \item If positive, append counter to the array and set it back to zero.
        \item \Red{Increment counter by 1 before moving on to the next vertex.}
    \end{itemize}
\end{enumerate}
In Figure~\ref{fig:series1} we show an example network and how the RS is extracted from it by following the algorithm described above.

\begin{figure*}[htbp]
    \centering
    \includegraphics[scale=0.45]{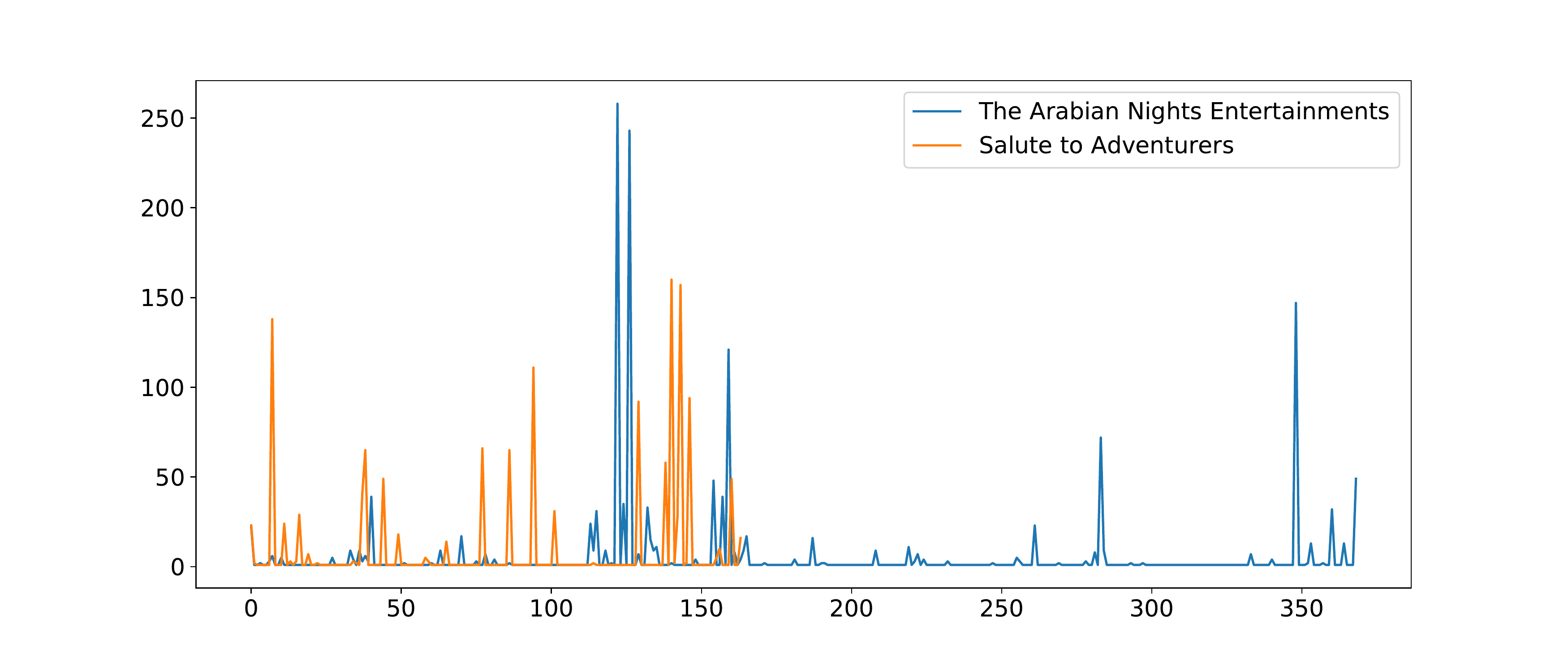}
    \caption{Plots of the recurrence signatures proposed for two books: ``The Arabian Nights Entertainments" in blue and ``Salute to Adventurers" in orange.}
    \label{fig:recurrence-signatures}
\end{figure*}

This signature is capable of representing properties of the network itself and, essentially, the overall story flow told by the book in question. In Figure \ref{fig:recurrence-signatures}, in blue, one can see the signature plot for the book \textit{The Arabian Nights Entertainments} and for \textit{Salute to Adventurers}, in orange. In these plots, the x axis is simply the index of the value in the number series, while the y axis is the value itself. With these illustrations, it is possible to observe and compare some specific properties of each narrative, such as the frequency of the story line bents, the length of these connections and how long it takes for another one to happen. Therefore, this series can be used to represent a specific network and, moreover, a specific book. Hence, this measure can potentially be used to solve different tasks, e.g. genre prediction, which will be discussed in the following section.

\section{\label{sec:model} Results and discussion}

\subsection{Discrimination between real and meaningless texts}

The first experiment performed was meant to evaluate how well the proposed methodology was able to grasp the sense of narrative from a text. For that, we extended the dataset by considering its shuffled version for each book. The shuffled version of a book is obtained by randomizing the order of the paragraphs of the text without modifying anything inside any of the paragraphs. Therefore, even though we maintain each sentence's syntactical and semantic structure unharmed, the sense of narrative for the whole text is lost since there is no meaningful story line anymore.
Then, to discriminate between real and meaningless texts, we considered three different measurements extracted from the mesoscopic network: 
\begin{itemize}
    \item mean of the accessibility for the second concentric level of every node: $mean_k = mean(\{k_1(2), k_2(2), ..., k_n(2)\}$)
    \item standard deviation of the accessibility for the second concentric level of every node: $std_k = std(\{k_1(2), k_2(2), ..., k_n(2)\}$)
    \item mean of the backbone symmetry for the second concentric level of every node: $mean_S = mean(\{S_1(2), S_2(2), ..., S_n(2)\}$)
\end{itemize}

\Red{Since the network's topological characteristics are similar to the small world model, considering too high concentric level values would cause an overall sweep of the network at once, and it would not be possible to properly discriminate the nodes within the network. Therefore, smaller concentric levels are a better fit to the problem and, after empirical tests, $h=2$ yielded the most satisfactory results in the experiments.}

\begin{figure}[htbp]
    \centering
    \includegraphics[scale=0.5]{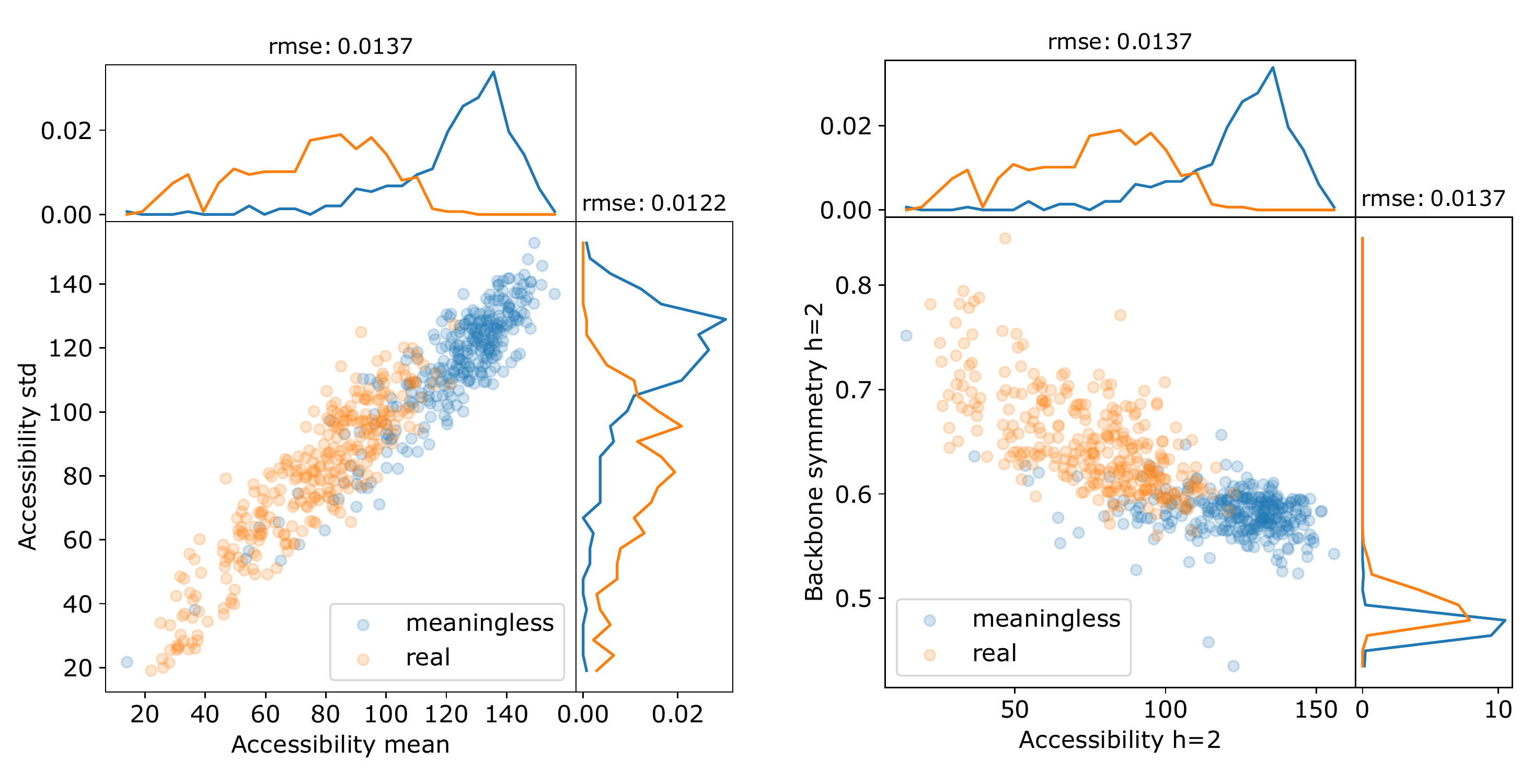}
    \caption{Discriminating between real texts and meaningless ones using accessibility's mean and standard deviation extracted from the mesoscopic network, on the left. On the right, the same discrimination task, but now using accessibility and backbone symmetry means. Both cases considering $h=2$ for the measurements calculation.}
    \label{fig:result_random}
\end{figure}

Figure~\ref{fig:result_random}, on the left, displays the obtained results when considering $mean_k$ and $std_k$ to discriminate between real (in orange) and meaningless (in blue) texts. It is visually noticeable that the two categories are separately clustered in the plot, indicating that our approach can successfully discriminate between them. We also consider the Root Mean Squared Error (RMSE) between the real and meaningless clusters to quantify how much each considered measurement contributes to their separation.

Additionally, the plot on the right of Figure~\ref{fig:result_random} displays the obtained results when considering $mean_k$ and $mean_S$ to discriminate between real (in orange) and meaningless (in blue) texts. Again, by observing the plot and evaluating the RMSE obtained for the aforementioned measurements, it is possible to conclude that this set of measures can also successfully discriminate between real and meaningless texts, therefore indicating that the measurements extracted from the mesoscopic network can, indeed, capture the narrative of the text.

\subsection{Genre discrimination}
The second experiment performed aimed at verifying how much the proposed methodology was capable of discriminating books respectively to literary genres. To make that possible, we labeled the dataset according to the genre communities, which yielded two different groups: \textit{fiction} and \textit{others} (see Figure~\ref{fig:communities}). 
As described in Section \ref{sec:dataset}, there are three main communities in the genre network: \textit{Adult Fiction}, \textit{Juvenile Fiction} and \textit{Others}. Here, we make use of these communities and the genres set provided by the Gutenberg Project to define one specific label for each book, regarding its literary genre. For a book to be labeled into the \textit{fiction} group, the majority of its genres listed in the Gutenberg dataset must belong to either of the fiction groups, otherwise the book will be labeled \textit{others}. 
For this task, we also considered the same three measurements mentioned earlier: $mean_k$, $std_k$ and $mean_s$.

\begin{figure*}[htbp]
    \centering
    \includegraphics[scale=0.5]{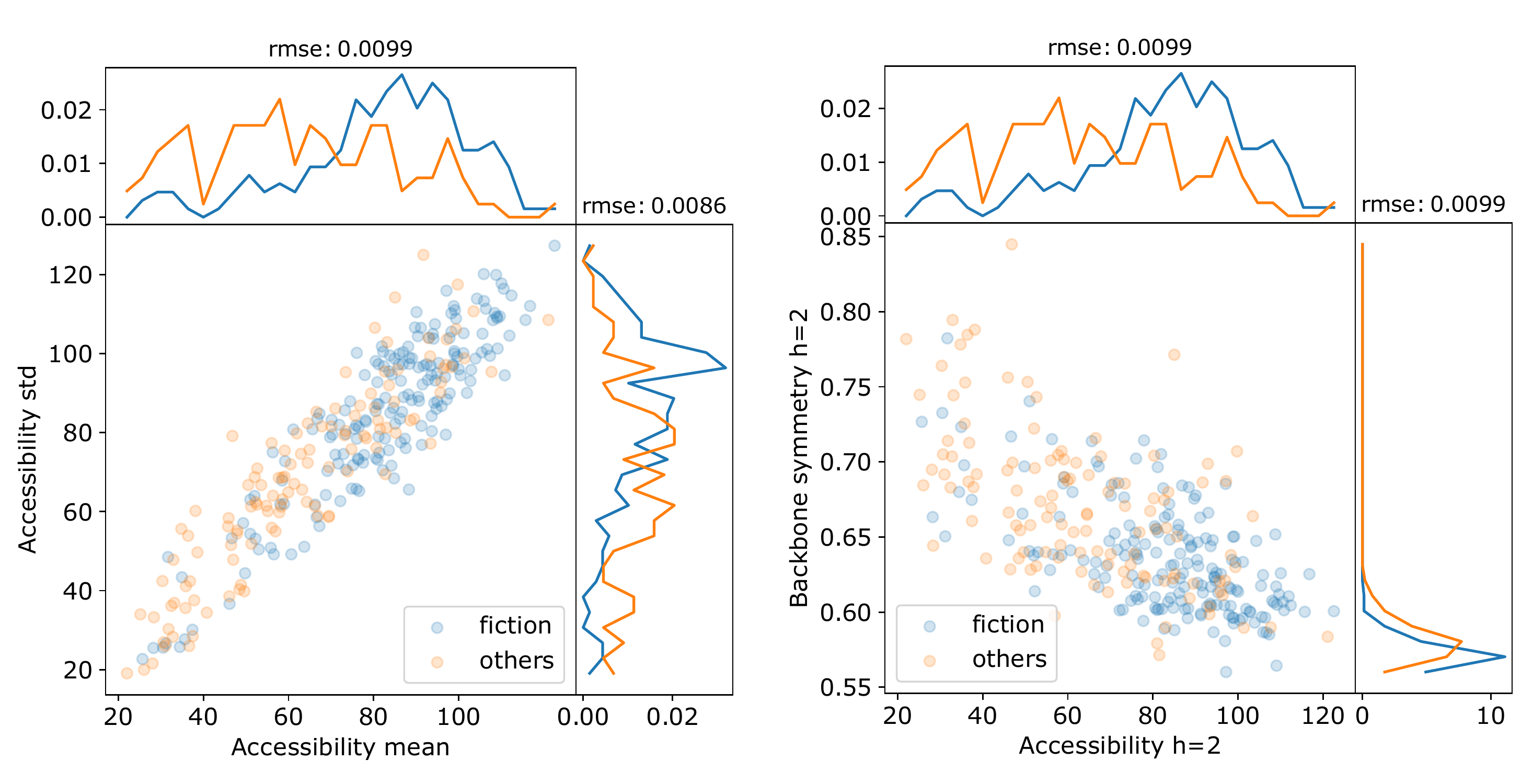}
    \caption{Discriminating literary genres using accessibility's mean and standard deviation extracted from the mesoscopic network, on the left. On the right, the same discrimination task, but now using accessibility and backbone symmetry means. Both cases considering $h=2$ for the measurements calculation.}
    \label{fig:result-genre}
\end{figure*}

The chart on the left of Figure~\ref{fig:result-genre} displays the obtained results when considering $mean_k$ and $std_k$ to discriminate between \textit{fiction} (in orange) and \textit{others} (in blue). Even though the separation of the groups is not trivially observed in the chart as before, the RMSE values obtained are sufficiently high to indicate that the measures are considerably discriminating between the two groups. Furthermore, density plots for each of the measurements also point to a significant distribution difference between them. This same discriminability of literary genres can not be obtained with a tf-idf approach alone, as discussed in Appendix B. 

Additionally, the chart on the right of Figure~\ref{fig:result-genre} displays the obtained results when considering $mean_k$ and $mean_S$ to discriminate between \textit{fiction} (in orange) and \textit{others} (in blue). Again, by observing the plots and evaluating the RMSE values obtained, it is also possible to notice a tendency to distinguish the two literary genres in question. This fact indicates that the genre label is somehow reflected in the narrative constructed in a book. More specifically, texts within the \textit{fiction} group are written in a similar way, with a similar story flow, whereas the books belonging to the \textit{others} group are slightly different from them, according to our methodology.

\subsection{Contributions of the recurrence signature}

Since the Recurrence Signature was developed in the context of text characterization, it has a more straightforward interpretation when portraying some of a book's specific properties. By definition, the RS represents the unfolding of the story flow in a book, by considering the chronological facts told and also any reference to another part of the book.   This motivation for this proposed measure suggests that it can be particularly useful in the context of text characterization.

\begin{figure*}[htbp]
    \centering
    \includegraphics[scale=0.55]{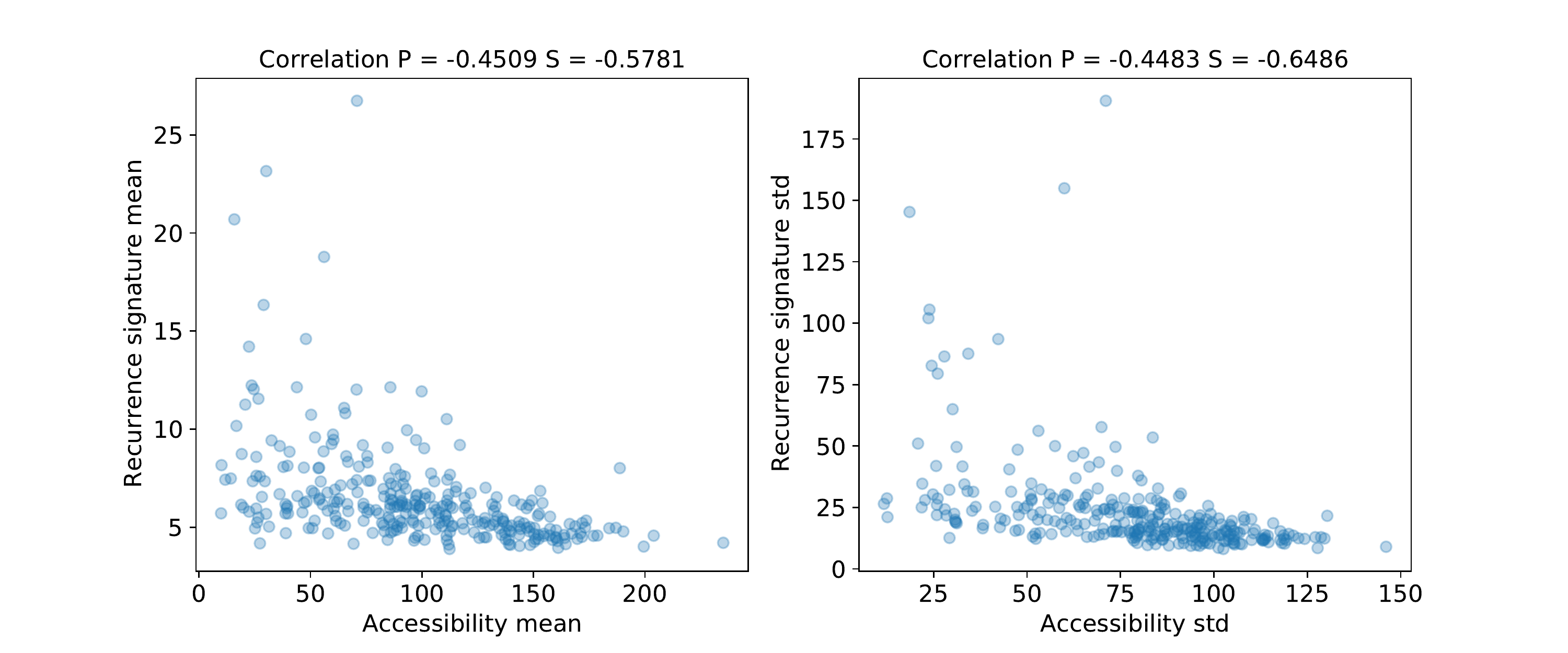}
    \caption{On the left, the chart illustrates the correlation between the mean of accessibility (considering $h=3$) and the mean of RS. On the right, the correlation between the standard deviation of accessibility (considering $h=3$) and the standard deviation of RS.}
    \label{fig:correlation}
\end{figure*}

One interesting result observed regarding the recurrence signature proposed is its relationship to the accessibility measure~\cite{TRAVENCOLO200889}. When evaluating the correlation between the mean of RS and standard deviation to other known network measurements (e.g., number of nodes, clustering coefficient, accessibility, etc.), we came across relatively high correlation values for the pairs involving the accessibility measure. This relationship is shown in the scatter plots in Figure~\ref{fig:correlation}. The chart on the left displays a considerable correlation between the mean of RS and the accessibility mean, which yields a Pearson correlation value of $-0.45$ and a Spearman value of $-0.58$. Likewise, the chart on the right illustrates the similar relationship between the standard deviation of RS and the standard deviation of accessibility, which yielded Pearson and Spearman correlation values of $-0.45$ and $-0.65$, respectively.

This relationship between the RS and the accessibility is interesting since it indicates that they both capture similar characteristics of a network's topology. The accessibility uses the idea of random walks to analyze the most likely paths within the network while considering each node's centrality or periphery. By considering a distance $h=3$, for example, this measurement agrees with the RS results, possibly reflecting the long-distance connections between distant nodes in the network or, in other words, parts of the book. It is also interesting to note the importance of selecting an appropriate value for $h$, when considering lower values such as $2$ or $3$, the results obtained were more satisfactory than for the larger ones. This happens because of the Small World-like topology of the networks considered, that contributes to an almost complete traversal coverage of the network in just a few steps. Hence, when considering values such as $4$ or $5$ for $h$, almost the whole network can be reached from any node, and then the results obtained are not as useful.

In this section we presented and discussed the capabilities of the accessibility and symmetry measures when applied to the mesoscopic network to discriminate between meaningful narratives and literary genres. It was also shown how the Recurrence Signature proposed can relate to the accessibility measurement. The considerable, yet not too high, correlation values encountered indicate that, even though both of them can capture some similar properties of the network, they are not redundant, which is of particular interest. The Recurrence Signature was developed to have a more straightforward interpretation, given the text characterization context. The results obtained indicate a good potential of the proposed measurement, that can successfully capture network properties related to the books narrative.

\section{\label{sec:conc} Conclusions}

Several interesting problems involving complex systems and structures are characterized by presenting properties along several scales.  In the case of literary texts, their properties extend from more microscopic characteristics such as the vowels and consonants composing words to more macroscopic organization of the text along successive chapters.    
The present work aimed at developing further some previous approaches focusing on \emph{mesoscopic} characterization of text~\cite{ferraz2018representation,DEARRUDA2018110}. Here, we aimed at investigating to what an extent different book genres can be discriminated from three main types of measurements: (a) accessibility~\cite{TRAVENCOLO200889}; (b) symmetry~\cite{SILVA201661}; and (c) recurrence signatures, the latter being proposed in the present work.

We considered 300 books from the Gutenberg project~\cite{gerlach2020standardized},
organized into two major genres.  First, the texts were pre-processed and segmented into paragraphs, and the co-reference resolution technique was applied. Next, we employ a syntactic analysis to select the most contextual meaning of the sentences. More specifically, the paragraphs were converted into specific syntactic parts (subject, verb, or direct object). Finally, the mesoscopic network was created by considering the cosine similarity between the paragraphs.
Two main types of experiments were performed, comparing
real texts with meaningless texts; and comparing fiction and others.

The obtained results indicate that both the accessibility and symmetry allowed reasonable separation in both experiments, indicating that this measurement was capable of capturing, to some extent, the heterogeneity along the narrative.  Corroborating the importance of scale in the analysis of texts, it has been verified that the best separations between the literary genres were obtained while choosing $h=2$ as the topological scale for the accessibility and symmetry measurements. 
The recurrence signatures were also capable of indicating differences between the considered types of text, presenting a moderate correlation with the accessibility, with potential for reflecting in a more direct manner aspects of the book narrative.

Among the related future developments we have the consideration of additional books and genres, as well as other complementary measurements including the BERT approach~\cite{devlin2018bert}.  In particular, it would be interesting to incorporate methods capable of capturing semantic aspects of the narrative.

\section*{Acknowledgments}
B. C. e Souza acknowledges CAPES for sponsorship - Brasil (CAPES) - Finance Code 001. 
% B. C. e Souza achnowledges that this study was financed in part by the Coordenação de Aperfeiçoamento de Pessoal de Nível Superior - Brasil (CAPES) - Finance Code 001"
H. F. de Arruda acknowledges FAPESP for sponsorship (grant no. 2018/10489-0) and Soremartec S.A. and Soremartec Italia, Ferrero Group, for partial financial support (from 1st July 2021). His funders had no role in study design, data collection, and analysis, decision to publish, or manuscript preparation.
L. da F. Costa thanks CNPq (grant no.  307085/2018-0).
D. R. Amancio thanks FAPESP (grant no. 20/06271-0) and CNPq (grant no. 304026/2018-2 and 311074/2021-9). This work has been supported also by the FAPESP grant 15/22308-2 and 21/01744-0.

\newpage

\section*{Appendix A: Computational expenses} 

In order to have a better idea of the computational expenses involved in the adopted methodology, we separated the tasks into three main groups: (a) Text processing (including syntactic analysis and text cleanup); (b) Network modelling (including the calculation of the edge weights and their respective pruning); and (c) Network characterization (including the calculation and extraction of the accessibility and symmetry measures and the proposed recurrence signature).  

The whole procedure was executed ten times respectively to the longest book, and the obtained average and standard deviation of the execution times were, in seconds:
(a) $84.8 \pm 5.5$; (b) $151.4 \pm 20.8$; and (c) $0.5 \pm 0.2$, resulting in a total time of $236.7 \pm 26.1$ seconds. These results refer to execution in an x86-64 i7 1.8 Ghz CPU, with the applications being implemented in Python.

The longest execution time was observed for the network modeling tasks, followed by text processing, with the network characterization corresponding to a fraction of the other times.  Network modeling required the longest execution time because it requires the calculation of the cosine distance between all possible pairwise combinations of nodes (paragraph windows).  The network characterization tasks resulted noticeably fast, as it requires a relatively small hierarchical neighborhood.

\newpage

\section*{Appendix B: TF-IDF contribution alone}
\begin{figure*}[htbp]
    \centering
    \includegraphics[scale=0.55]{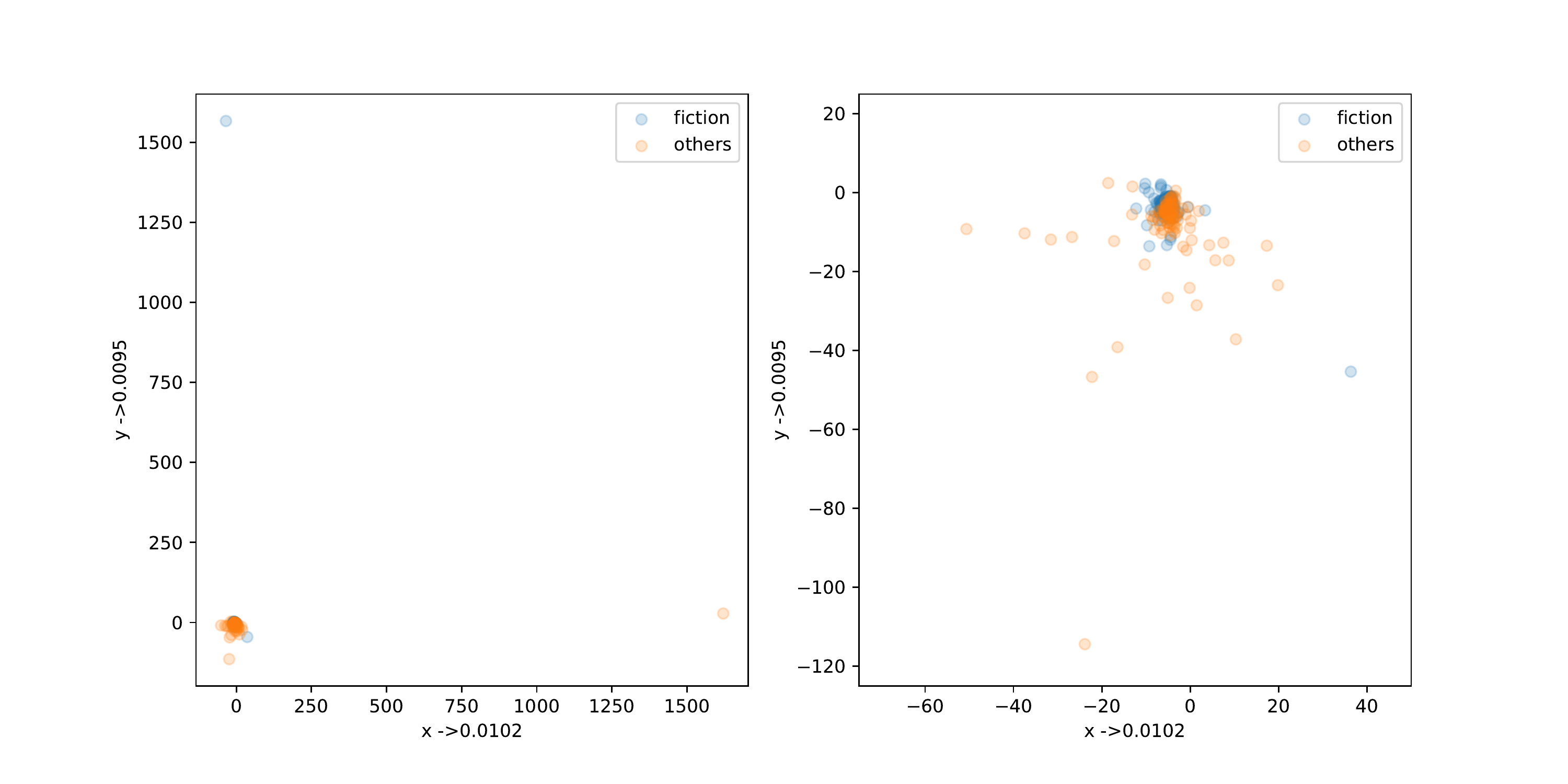}
    \caption{Scatter plot of what was obtained from employing the PCA technique onto the data provided by the TF-IDF, colored by the literary genre label of each book. On the left the original plot, and the cropped version on the right.}
    \label{fig:tfidf}
\end{figure*}

To ensure that the obtained results were not coming from the employment of the TF-IDF alone, we performed the following experiment. By considering the TF-IDF vectors obtained from the books as features for the PCA technique, we generated the scatter plots on Figure~\ref{fig:tfidf}. As one can see, there was no visual clustering evident in the results, indicating that the TF-IDF alone is not sufficient to discriminate between the different books.

\bibliographystyle{abbrvnat}
%\bibliography{references}

\end{document}